# Optimization of Evolutionary Neural Networks Using Hybrid Learning Algorithms

Ajith Abraham

School of Business Systems, Monash University, Clayton, Victoria 3800, Australia, Email: ajith.abraham@ieee.org

**Abstract** - Evolutionary artificial neural networks (EANNs) refer to a special class of artificial neural networks (ANNs) in which evolution is another fundamental form of adaptation in addition to learning. Evolutionary algorithms are used to adapt the connection weights, network architecture and learning algorithms according to the problem environment. Even though evolutionary algorithms are well known as efficient global search algorithms, very often they miss the best local solutions in the complex solution space. In this paper, we propose a hybrid meta-heuristic learning approach combining evolutionary learning and local search methods (using $1^{st}$ and $2^{nd}$ order error information) to improve the learning and faster convergence obtained using a direct evolutionary approach. The proposed technique is tested on three different chaotic time series and the test results are compared with some popular neuro-fuzzy systems and a recently developed cutting angle method of global optimization. Empirical results reveal that the proposed technique is efficient in spite of the computational complexity.

## I. INTRODUCTION

At present, neural network design relies heavily on human experts who have sufficient knowledge about the different aspects of the network and the problem domain. As the complexity of the problem domain increases, manual design becomes more difficult and unmanageable. The interest in evolutionary search procedures for designing ANN architecture has been growing in recent years as they can evolve towards the optimal architecture without outside interference, thus eliminating the tedious trial and error work of manually finding an optimal network [2] [4] [18]. The advantage of the automatic design over the manual design becomes clearer as the complexity of ANNs increases [1]. EANNs provide a general framework for investigating various aspects of simulated evolution and learning. In EANN's evolution can be introduced at various levels. At the lowest level, evolution can be introduced into weight training, where ANN weights are evolved. At the next higher level, evolution can be introduced into neural network architecture adaptation, where the architecture (number of hidden layers, no of hidden neurons and node transfer functions) is evolved. At the highest level, evolution can be introduced into the learning mechanism [3] [18].

Global optimization is concerned with finding the best possible solution to a given problem. As there are no efficient algorithms to achieve this goal in general, heuristic global optimization methods like evolutionary algorithms are often used to optimize neural networks. Alternatively, it is sometimes acceptable to find a local optimum, which is as good as all solutions in its neighborhood. Local search methods are comparatively well understood, and local optima can often be found efficiently even for problems in which global optimization is difficult. Evolution of connection weights at the lowest level and the learning algorithm at the highest level (using $1^{st}$ and $2^{nd}$ order error information) could be considered as a hybrid learning approach wherein the initial weights determined by the evolutionary learning process is fine tuned by the local search method. The parameters of the local search algorithm could be optimized using the evolutionary search process. In Section II, we present the details of the proposed meta-heuristic learning approach followed by experimentation setup and results in Section III - V. Some discussions and conclusions are provided in Section VI and VII.

## II. EANN OPTIMIZATION

One major problem of evolutionary algorithm is their inefficiency in fine tune local search although they are good at global search. The efficiency of evolutionary training can be improved significantly by incorporating a local search procedure into the evolution. Evolutionary algorithms are used to first locate a good region in the space and then a local search procedure is used to find a near optimal solution in this region. It is interesting to consider finding good initial weights as locating a good region in the space. Defining that the basin of attraction of a local minimum is composed of all the points, sets of weights in this case, which can converge to the local minimum through a local search algorithm, then a global minimum can easily be found by the local search algorithm if the evolutionary algorithm can locate any point, i.e, a set of initial weights, in the basin of attraction of the global minimum. Referring to Figure 1, $G_1$ and $G_2$ could be considered as the initial weights as located by the evolutionary search and $W_A$ and $W_B$ the corresponding final weights fine-tuned by the meta-learning approach.

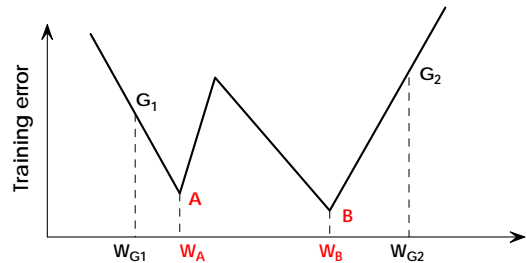

**Figure 1.** Fine tuning of weights using meta-learning

Figure 2 illustrates the general interaction mechanism with the learning mechanism of the EANN evolving at the highest level on the slowest time scale. All the randomly generated architecture of the initial population are trained by four different learning algorithms [9] (backpropagation - BP, scaled conjugate gradient - SCG, quasi-Newton algorithm -

QNA and Levenberg-Marquardt - LM) and evolved in a parallel environment. Parameters controlling the performance of the learning algorithm will be adapted (example, learning rate and momentum for BP) according to the problem. The basic algorithm of the proposed meta-heuristic approach is as follows:

1. *Set t=0 and randomly generate an initial population of neural networks with architectures, node transfer functions and connection weights assigned at random.*
2. *In a parallel mode, train each network using BP/SCG/QNA and LM for the specified number of iterations.*
3. *Based on fitness value, select parents for reproduction*
4. *Apply mutation to the parents and produce offspring (s) for next generation. Refill the population back to the defined size.*
5. *Repeat step 2*
6. *STOP when the required solution is found or number of iterations has reached the required limit.*

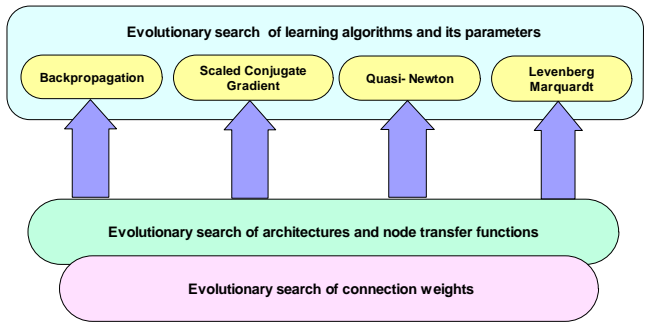

**Figure 2.** Interaction of various evolutionary search mechanisms

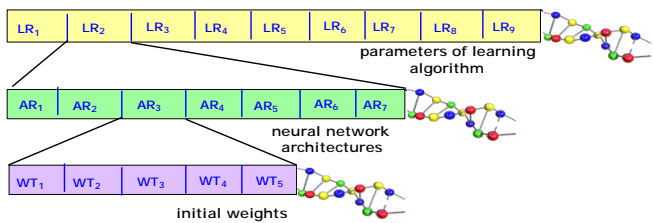

**Figure 3.** Chromosome representation

Architecture of the chromosome for the proposed hierarchical evolutionary search is depicted in Figure 3. From the point of view of engineering, the decision on the level of evolution depends on what kind of prior knowledge is available. If there is more prior knowledge about EANN's architectures than that about their learning rules or a particular class of architectures is pursued, it is better to implement the evolution of architectures at the highest level because such knowledge can be used to reduce the search space and the lower level evolution of learning algorithms can be more biased towards this kind of architectures. On the other hand, the evolution of learning algorithms should be at the highest level if there is more prior knowledge about them available or there is a special interest in certain type of learning algorithm.

Connection weights may be represented as binary strings represented by a certain length. The whole network is encoded by concatenation of all the connection weights of the network in the chromosome. A heuristic concerning the order of the concatenation is to put connection weights to the same node together. Evolutionary architecture adaptation can be achieved by constructive and destructive algorithms. Constructive algorithms, which add complexity to the network starting from a very simple architecture until the entire network is able to learn the task. Destructive algorithms start with large architectures and remove nodes and interconnections until the ANN is no longer able to perform its task. Then the last removal is undone. Direct encoding of the architecture makes the mapping simple but often suffers problems like scalability and implementation of crossover operators. For an optimal network, the required node transfer function (gaussian, sigmoidal, etc.) could be formulated as a global search problem, which is evolved simultaneously with the search for architectures. For the neural network to be fully optimal the learning algorithms are to be adapted dynamically according to its architecture and the given problem. For BP, deciding the optimal learning rate and momentum can be considered as the first attempt of adaptation of the local search technique (learning algorithm). The best learning algorithm will again be decided by the evolutionary search mechanism. Genotypes of the learning parameters of the different learning algorithms can be encoded as real-valued coefficients [4].

### III. DATA SETS FOR EXPERIMENTATION

In our experiments, we used the following 3 different time series for training the neural networks and evaluating the performance.

#### a) Waste Water Flow Prediction

The problem is to predict the wastewater flow into a sewage plant [13]. The data set is represented as [$f(t), f(t-1), a(t), b(t), f(t+1)$] where $f(t), f(t-1)$ and $f(t+1)$ are the water flows at time *t,t-1, and t+1* (hours) respectively. $a(t)$ and $b(t)$ are the moving averages for 12 hours and 24 hours. The time series consists of 475 data points. The first 240 data sets were used for training and remaining data for testing.

#### b) Mackey-Glass Chaotic Time Series

The Mackey-Glass differential equation is a chaotic time series [14] for some values of the parameters $x(0)$ and $\tau$.

$$\frac{dx(t)}{dt} = \frac{0.2x(t-\tau)}{1 + x^{10}(t-\tau)} - 0.1\,x(t)$$

We used the value $x(t-18), x(t-12), x(t-6), x(t)$ to predict $x(t+6)$. Fourth order Runge-Kutta method was used to generate 1000 data series. The time step used in the method is

0.1 and initial condition were $x(0)=1.2$, $\tau=17$, $x(t)=0$ for $t < 0$. First 500 data sets were used for training and remaining data for testing.

### c) Gas Furnace Time Series Data

This time series was used to predict the $CO_2$ (carbon dioxide) concentration $y(t+1)$ [7]. Data is represented as [$u(t)$, $y(t)$, $y(t+1)$] The time series consists of 292 pairs of observation and 50% of data was used for training and remaining for testing.

**TABLE 1. PARAMETERS USED FOR EANNs**

| Parameter | Setting |
| --- | --- |
| Population size | 40 |
| Maximum no of generations | 40 |
| Number of hidden nodes | (a) max. 16 neurons (b) max. 4 neurons |
| Activation functions | tanh (T), logistic (L), sigmoidal (S), tanh-sigmoidal (T*), log-sigmoidal (L*) |
| Output neuron | linear |
| Training epochs | 500 |
| Initialization of weights | +/- 0.3 |
| Ranked based selection | 0.50 |
| Elitism | 5 % |
| Mutation rate | 0.40 |

**TABLE 2. PARAMETERS OF THE LEARNING ALGORITHMS**

| Learning algorithm | Parameter | Setting |
| --- | --- | --- |
| Backpropagation | Learning rate | 0.25-0.05 |
|  | Momentum | 0.25-0.05 |
| Scaled conjugate gradient algorithm | Change in weight for second derivative approximation | 0 - 0.0001 |
|  | Regulating the indefiniteness of the Hessian | 0 – 1.0 E-06 |
| Quasi-Newton algorithm | Step lengths | 1.0E-06 – 100 |
|  | Limits on step sizes | 0.1 – 0.6 |
|  | Scale factor to determine performance | 0.001 – 0.003 |
|  | Scale factor to determine step size. | 0.1 - 0.4 |
| Levenberg Marquardt | Learning rate | 0.001 – 0.02 |

## IV. EXPERIMENTATION SETUP AND RESULTS

We have applied the proposed technique to the three-time series mentioned in Section III. For performance comparison, we also trained neural networks (pre-defined architecture, node transfer functions and random initialization of weights) using the same training and test data sets. The parameters used in our experiments were set to be the same for all the 3 data sets. Fitness value is calculated based on the RMSE achieved on the test set. In this experiment, we have considered the best-evolved neural network as the best individual of the last generation. All the genotypes were represented using binary coding and the initial populations were randomly created based on the parameters shown in Table 1. The parameter settings, which were evolved for the different learning algorithms, are illustrated in Table 2. The experiments were repeated three times and the worst RMSE values are reported. Please refer to Table 3 for empirical values of RMSE on training and test data for the three time series problems using the hybrid learning technique. The performance of the technique was evaluated with (a) maximum number of 16 hidden neurons and (b) maximum number 4 hidden neurons. For comparison purposes, test set RMSE values for neural networks using conventional design techniques (using maximum 24 hidden neurons) are also presented in Table 3. Figures 4 - 6 depict the convergence of the meta-learning approach after 40 generations. Figures 7 - 9 illustrates the desired and predicted values of the different time series using BP learning.

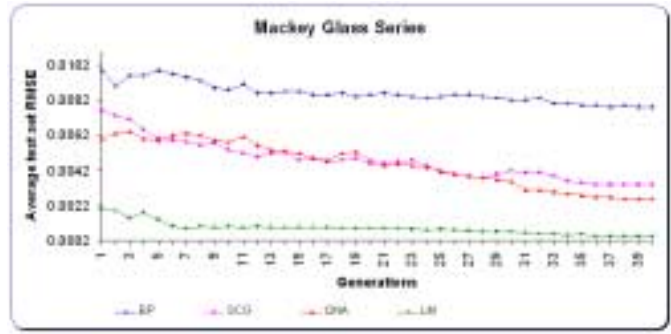

**Figure 4.** Mackey Glass time series: Average test set RSME

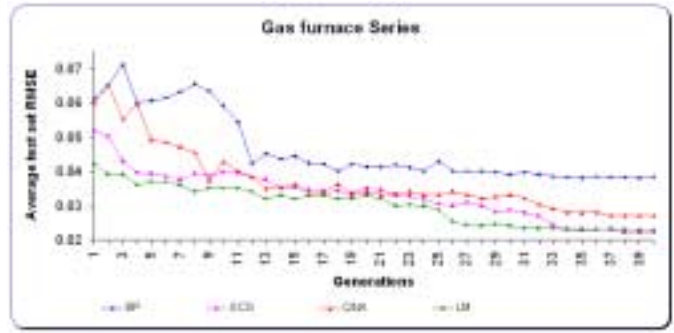

**Figure 5.** Gas furnace time series: Average test set RSME

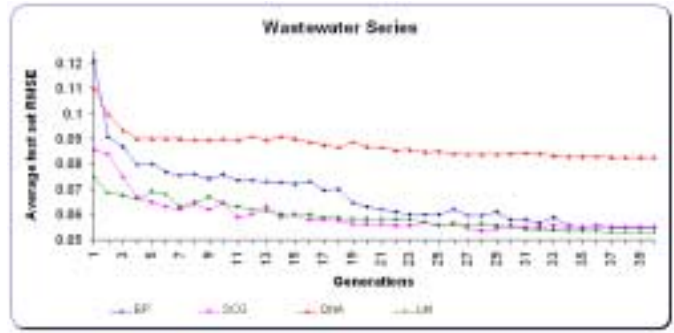

**Figure 6.** Wastewater time series: Average test set RSME

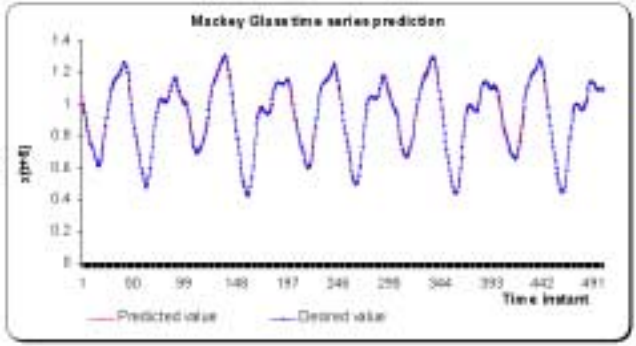

**Figure 7.** Mackey Glass series: Test results for BP hybrid learning

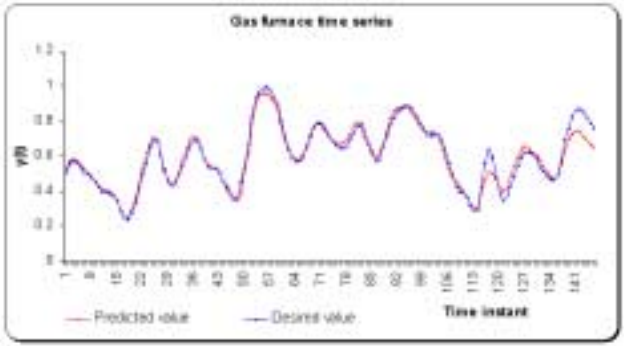

**Figure 8.** Gas furnace series: Test results for BP hybrid learning

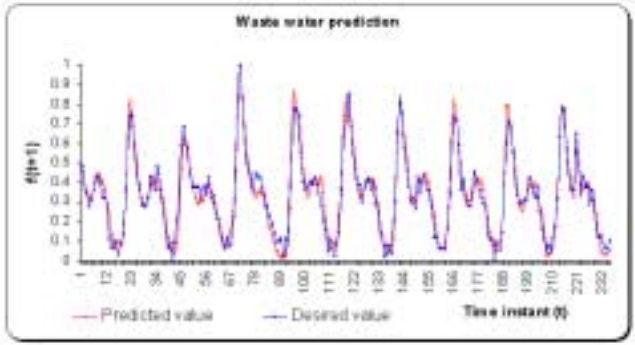

**Figure 9.** Waste water series: Test results for BP hybrid learning

## V. COMPARISON WITH OTHER TECHNIQUES

In this section we explore the performance of the proposed algorithm with the following intelligent techniques.

### Neuro-fuzzy systems

Neuro-fuzzy (NF) computing is a popular framework for solving complex problems [1]. If we have knowledge expressed in linguistic rules, we can build a Fuzzy Inference System (FIS) [8], and if we have data, or can learn from a simulation (training) then we can use ANNs. For building a FIS, we have to specify the fuzzy sets, fuzzy operators and the knowledge base. Similarly for constructing an ANN for an application the user needs to specify the architecture and learning algorithm. An analysis reveals that the drawbacks pertaining to these approaches seem complementary and therefore it is natural to consider building an integrated system combining the concepts. While the learning capability is an advantage from the viewpoint of FIS, the formation of linguistic rule base will be advantage from the viewpoint of ANN. The neuro-fuzzy models considered were Dynamic Evolving Fuzzy neural networks (dmEFuNN) [12] implementing a Mamdani FIS [15] and an Adaptive Neuro-Fuzzy Inference System (ANFIS) [11] implementing a Takagi-Sugeno FIS [16]. The same training and test sets of the three time series were used to compare the performance with the neuro-fuzzy systems. The empirical results are depicted in Table 4.

### ANN Optimization Using Cutting Angle Method (CAM)

The cutting angle method is based on theoretical results in abstract convexity [5]. It systematically explores the whole domain by calculating the values of the objective function $f(\mathbf{x})$ at certain points. The points are selected in such a way that the algorithm does not return to unpromising regions where function values are high. The new point is chosen where the objective function can potentially take the lowest value. The function is assumed to be Lipschitz, and the value of the potential minima is calculated based on both the distance to the neighbouring points and function values at these points. This process can be seen as constructing the piecewise linear lower approximation of the objective function $f(x)$. With the addition of new points, the approximation $h_k(x)$ becomes closer to the objective function, and the global minimum of the approximating function x* converges to the global minimum of the objective function. The lower approximation, the auxiliary function $h_k(x)$, is called the saw-tooth cover of $f$. The architecture of the ANN was prefixed and after the specified number of CAM iterations (2000 for waste water, 3000 for Mackey Glass and 1000 for Gas furnace series), the error was further improved using 300 iterations of Levenberg-Marquardt algorithm. The comparative performance with hybrid learning technique is depicted in Table 5 [6].

## VI. DISCUSSIONS

Table 3 shows comparative performance between the hybrid optimization technique and a conventional ANN with a maximum of 24 hidden neurons (2500 epochs training). We also compared the performance when the search space of the number of hidden neurons was decreased from 16 to 4 nos. LM algorithm produced best results with 4 hidden neurons for all the 3 data sets. However, when the hidden neurons were increased, SCG algorithm marginally preformed better than LM. For Mackey glass series the results were not that encouraging (using 4 hidden neurons) when compared with the conventional design using 24 hidden neurons. As depicted in Table 3, Mackey Glass series requires more hidden neurons to improve the RMSE values. Table 4 and 5 depicts empirical comparison between the hybrid learning approach, neuro-fuzzy systems and CAM. As evident, the hybrid learning approach has outperformed all the intelligent techniques in terms of the lowest RMSE values on test set for the time series considered.

**TABLE 3. PERFORMANCE COMPARISON BETWEEN HYBRID LEARNING APPROACH AND CONVENTIONAL NEURAL NETWORKS**

| Data set | Local search algorithm | Hybrid EANN (maximum 16 hidden neurons) | | | Hybrid EANN (maximum 4 hidden neurons) | | ANN (2500 epochs training) | |
|---|---|---|---|---|---|---|---|---|
| | | RMSE Training | RMSE Test | Architecture | RMSE Training | RMSE Test | RMSE Test | Architecture |
| Mackey Glass | BP | 0.0072 | 0.0077 | 7 T, 3 L | 0.0166 | 0.0168 | 0.0437 | 24 T* |
| | SCG | 0.0030 | 0.0031 | 11 T | 0.0062 | 0.0067 | 0.0045 | 24 T* |
| | QNA | 0.0024 | 0.0027 | 6 T, 4 T* | 0.0059 | 0.0058 | 0.0034 | 24 T* |
| | LM | 0.0004 | †0.0004 | 8 T, 2 T*, 1 L* | 0.0056 | †0.0061 | 0.0009 | 24 T* |
| Gas Furnace | BP | 0.0159 | 0.0358 | 8 T | 0.0189 | 0.0371 | 0.0766 | 18 T* |
| | SCG | 0.0110 | †0.0210 | 8 T, 2 T* | 0.0179 | 0.0295 | 0.0330 | 16 T* |
| | QNA | 0.0115 | 0.0256 | 7 T, 2 L* | 0.0156 | 0.0295 | 0.0376 | 18 T* |
| | LM | 0.0120 | 0.0223 | 6 T, 1 L, 1 T* | 0.0181 | †0.0290 | 0.0451 | 14 T* |
| Waste Water | BP | 0.0441 | 0.0547 | 6 T, 5 T*,1 L | 0.0647 | 0.0639 | 0.1360 | 16 T* |
| | SCG | 0.0457 | 0.0579 | 6 T, 4 L* | 0.0580 | 0.0600 | 0.0820 | 14 T* |
| | QNA | 0.0673 | 0.0823 | 5 T, 5 TS | 0.0590 | 0.0596 | 0.1276 | 14 T* |
| | LM | 0.0425 | †0.0521 | 8 T, 1 LS | 0.0567 | †0.0591 | 0.0951 | 14 T* |

† Lowest RMSE value

**TABLE 4. PERFORMANCE COMPARISON BETWEEN HYBRID LEARNING APPROACH AND NEURO-FUZZY SYSTEMS**

| Data set | RMSE | | | | | |
|---|---|---|---|---|---|---|
| | Hybrid EANN (Best results adapted from Table 3) | | Mamdani - NF | | Takagi Sugeno - NF | |
| | Training | Test | Training | Test | Training | Test |
| Mackey Glass | 0.0004 | 0.0004 | 0.0023 | 0.0042 | 0.0019 | 0.0018 |
| Gas Furnace | 0.0110 | 0.0210 | 0.0140 | 0.0490 | 0.0137 | 0.0570 |
| Waste Water | 0.0425 | 0.0521 | 0.0019 | 0.0750 | 0.0530 | 0.0810 |

**TABLE 5. PERFORMANCE COMPARISON BETWEEN HYBRID LEARNING APPROACH AND CAM**

| Data set | Hybrid EANN (fine tuning by 400 epochs of LM) | | | CAM | | |
|---|---|---|---|---|---|---|
| | RMSE (train) | RMSE (test) | Architecture | RMSE (train) | RMSE (test) | Architecture |
| Mackey-Glass | 0.0056 | 0.0061 | 2 T, 2 T* | 0.0085 | 0.0091 | 4 S |
| Gas Furnace | 0.0181 | 0.0290 | 1 T, I L, 1 T* | 0.0173 | 0.0384 | 3 S |
| Waste water | 0.0567 | 0.0591 | 2 L, 1 T, 1 T* | 0.0570 | 0.0660 | 4 S |

## VII. CONCLUSIONS

Selection of the architecture (number of layers, hidden neurons, activation functions and connection weights) of a network and correct learning algorithm is a tedious task for designing an optimal artificial neural network. Moreover, for critical applications and hardware implementations optimal design often becomes a necessity. In this paper, we have formulated and explored; an adaptive computational framework based on evolutionary computation and optimization techniques for the automatic design of optimal artificial neural networks. Empirical results are promising and show the importance and efficacy of the technique. Different learning algorithms have their staunch proponents, who can always construct instances in which their algorithm performs better than most others. This study also reveals the difficulty to generalize which is the best local search algorithm that would work for all the problems. As evident from Table 3, for smaller networks, LM approach gave the best results.

Neuro-fuzzy system is able to precisely model the uncertainty and imprecision within the data as well as to incorporate the learning ability of neural networks. Even though the performance of neuro-fuzzy systems is dependent on the problem domain, very often the results are better while compared to pure neural network approach using gradient descent algorithm [1]. As depicted in Table 4, it is clear that an optimized neural network could easily outperform a neuro-fuzzy system. Compared to neural networks, an important advantage of neuro-fuzzy systems is its reasoning ability (*if-then* rules) of any particular state. A fully trained neuro-fuzzy system could be replaced by a set of *if-then* rules.

CAM technique optimizes the connection weights in a pre-defined architecture depending upon the designer's knowledge of the problem domain. Being deterministic, CAM uses a systematic exploration of the search space and it works best for small dimensional problems. As depicted in Table 5, CAM produced near optimal results when compared to evolutionary approach.

In the hybrid learning approach, our work was mostly concentrated on the evolutionary search of optimal learning algorithms and its parameters. For the evolutionary search of architectures, it will be interesting to model as co-evolving sub-networks [17] instead of evolving the whole network. Further, it will be worthwhile to explore the whole population information of the final generation for deciding the best solution [19]. We used a fixed chromosome structure (direct encoding technique) to represent the connection weights, architecture, learning algorithms and its parameters. As size of the network increases, the chromosome size grows. Moreover, implementation of crossover is often difficult due to production of non-functional offspring's. Parameterized encoding overcomes the problems with direct encoding but the search of architectures is restricted to layers. In the grammatical encoding rewriting grammar is encoded. So the success will depend on the coding of grammar (rules). Cellular configuration might be helpful to explore the architecture of neural networks more efficiently. Gutierrez et al [10] has shown that their cellular automata technique performed better than direct coding.